\documentclass[10pt,twocolumn,letterpaper]{article}

\usepackage{iccv}
\usepackage{times}
\usepackage{epsfig}
\usepackage{graphicx}
\usepackage{amsmath}
\usepackage{amssymb}
\usepackage{booktabs}
\usepackage{multirow}
\usepackage{float}
\usepackage[utf8]{inputenc}
\usepackage[accsupp]{axessibility}

\newcommand{\layoutbranch}{$\Phi_{layout}$\,}
\newcommand{\imagebranch}{$\Phi_{image}$\,}

\DeclareUnicodeCharacter{2744}{\snowflake}

\usepackage[pagebackref=true,breaklinks=true,letterpaper=true,colorlinks,bookmarks=false]{hyperref}

\iccvfinalcopy 

\def\iccvPaperID{3420} 

\ificcvfinal\fi

\begin{document}

\title{LaLaLoc: Latent Layout Localisation in Dynamic, Unvisited Environments}

\author{Henry Howard-Jenkins\\
Active Vision Laboratory\\
University of Oxford\\
{\tt\small henryhj@robots.ox.ac.uk}
\and
Jose-Raul Ruiz-Sarmiento\\
Machine Perception and\\ Intelligent Robotics Group\\
University of M\'{a}laga\\
{\tt\small jotaraul@uma.es}
\and
Victor Adrian Prisacariu\\
Active Vision Laboratory\\
University of Oxford\\
{\tt\small victor@robots.ox.ac.uk}
}

\maketitle
\ificcvfinal\thispagestyle{empty}\fi

\begin{figure*}[h]
    \centering
    \includegraphics[width=0.99\linewidth]{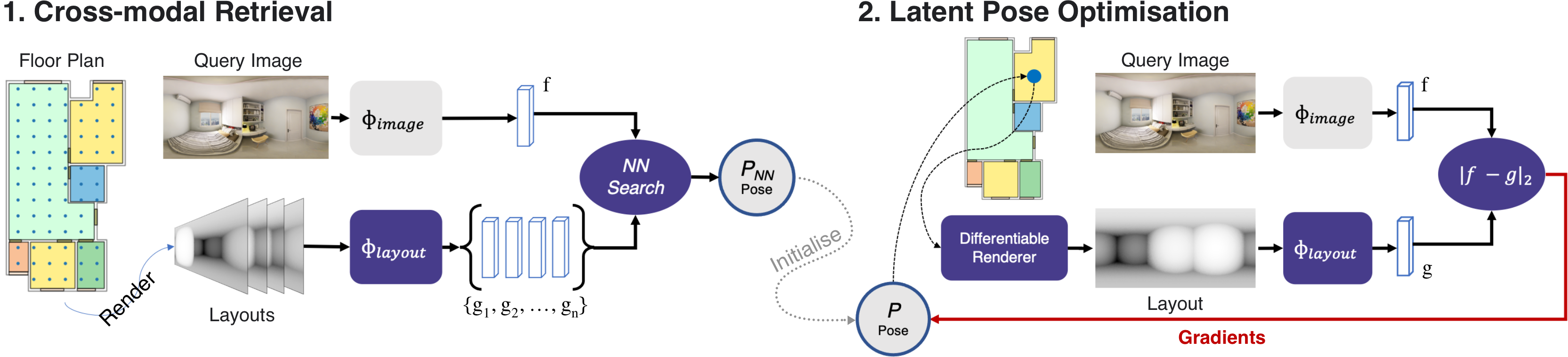}
    \caption{Overview of localisation using LaLaLoc. In the retrieval stage, the query image is mapped to the layout latent space by \imagebranch. We then sample a grid of poses from a known floor plan, render their layouts, and compute their respective latent representation through \layoutbranch. An initial pose estimate is found by a nearest neighbour search within this shared latent space. The nearest neighbour pose is then used as an initialisation for direct pose optimisation with our proposed latent pose optimisation. This is a gradient-based optimisation of pose conducted in the shared latent space, thus removing the need to decode into a common data mode. Our Vogel Disc resampling stage is ommitted from this viusalisation for clarity.}
    \label{fig:teaser}
\end{figure*}

\begin{abstract}

We present LaLaLoc to localise in environments without the need for prior visitation, and in a manner that is robust to large changes in scene appearance, such as a full rearrangement of furniture. Specifically, LaLaLoc performs localisation through latent representations of room layout. LaLaLoc learns a rich embedding space shared between RGB panoramas and layouts inferred from a known floor plan that encodes the structural similarity between locations. Further, LaLaLoc introduces direct, cross-modal pose optimisation in its latent space. Thus, LaLaLoc enables fine-grained pose estimation in a scene without the need for prior visitation, as well as being robust to dynamics, such as a change in furniture configuration. We show that in a domestic environment LaLaLoc is able to accurately localise a single RGB panorama image to within 8.3cm, given only a floor plan as a prior.
\end{abstract}


\section{Introduction}

Camera relocalisation is a fundamental problem in computer vision. Image-based relocalisation represents the goal of estimating the camera pose of an unseen image, given some prior knowledge about the surrounding environment. In this paper, we tackle the task of localisation in an environment that has not been previously visited, and one in which there may be considerable scene dynamics -- an area where significant scope for improvement has been identified~\cite{wald2020beyond}.

To address this, we propose to localise with respect to a known floor plan and the layout visible at a location within the scene. Floor plan-based localisation is particularly suited for the long-term localisation setting as, while objects and furniture may have moved, items represented in a structural floor plan, such as walls, floors and ceilings, will remain static. Therefore, it enables localisation over a long period of time without requiring continual re-training or re-mapping. In addition, in this formulation we only need the floor plan as prior, thus removing the need for previous visitation of the target environment, \textit{i.e.} without a training trajectory of images.

We present LaLaLoc, which performs floor plan-based localisation through latent representations of room layout. This layout latent space is cross-modal, shared between layouts inferred from the floor plan and the RGB panoramas queried at inference time. More specifically LaLaLoc performs localisation in two stages, depicted in Figure~\ref{fig:teaser}. The first stage provides a coarse estimate of pose through cross-modal retrieval. For the second stage, we propose a cross-modal direct optimisation of pose through differentiable rendering.

Differentiable rendering has been shown to be effective for object pose estimation~\cite{palazzi2018end, chen2019learning}. But these works typically rely on like-for-like rendering losses, such as the pixelwise error between the rendered and target images. However, since LaLaLoc operates across multiple modes of data between query and prior,
the prediction of a common data mode would be required for the comparison losses. Instead, we propose to optimise for pose directly in the layout latent space. Through this formulation, LaLaLoc is able to accurately align the floor plan to a cluttered RGB panorama without ever explicitly predicting its layout.

The contributions of this paper can be summarised as:
\begin{itemize}
    \item We propose LaLaLoc, a highly accurate localisation method that is robust to scene dynamics such as the configuration of furniture, and able to localise in a new scene without prior visitation.
    \item We introduce direct pose optimisation in the latent space. This allows for cross-modal pose optimisation, without the need for a decoder to traverse data modes for the computation of the matching cost.
    \item Through experimental evaluation, we demonstrate the accuracy of LaLaLoc as well as validate its formulation. This includes showing that the representation of room layout has significant influence on the efficacy of layout-based localisation and cannot be used interchangeably.
\end{itemize}

\section{Related Work}
A wide variety of methods have been produced to tackle the task of camera localisation. There are scene-specific methods, which require fine-tuning to each individual scene. Pose-regression methods~\cite{kendall2015posenet, kendall2017geometric, brahmbhatt2018geometry} train a deep network for each scene to directly predict the camera pose from an input, but these methods are limited in accuracy~\cite{sattler2019understanding}. Instead of regressing camera pose, scene-coordinate regression methods~\cite{shotton2013scene, brachmann2017dsac, brachmann2018learning, brachmann2019expert, valentin2015exploiting} densely predict 2D-3D correspondences between the query and the environment, which then allows for solving for pose via PnP. On the other hand, some groups of methods are able to generalise across scenes without need for retraining. Image retrieval methods~\cite{schindler2007city, arandjelovic2016netvlad, balntas2018relocnet,ding2019camnet} estimate pose through by using the pose of the most similar image within an image database. 3D structure-based methods~\cite{liu2017efficient,lim2012real,sarlin2019coarse} instead establish 2D-3D between the query image and points in a SfM model. However, the scene-specific and generalisable methods discussed do have a commonality in that they require prior visitation in the same data modality of the region in which localisation is performed: scene-specific methods require this for re-training, while the generalisable methods require it for map/database building. By contrast, LaLaLoc performs localisation without prior visitation, instead leveraging only a known floor plan as a prior.

Particularly within the field of robotics, there have been a few proposed approaches to localise with respect to a floor plan. These methods operate by aggregating depth~\cite{winterhalter2015accurate}, detecting suitable features such as layout corners~\cite{hile2008positioning}, or extracting layout edges~\cite{boniardi2019robot, wang2015lost, unicomb2018monocular} within query RGB and/or depth images to estimate and compare with the inferred layout at a location within the floor plan to estimate the observation likelihood.
When depth information is available, scan-matching techniques~\cite{pomerleau2015review} find a rigid alignment between the scene and the scan through alignment methods such as Iterative Closest Point (ICP)~\cite{besl1992icp}, General-ICP (GICP)~\cite{segal2009generalized}, or Normal Distributions Transform (NDT)~\cite{biber2003ndt}. ICP-based scan matching techniques have successfully been used to perform localisation within a floor plan~\cite{blum2020precise,watanabe2020robust}. These methods generally rely on a flow of information over multiple sequential measurements, typically alongside motion priors, such as from wheel or camera odometry, and are fed as hypothesis weightings into a Monte Carlo Localisation (MCL)~\cite{thrun2002probabilistic} framework. In this paper, however, we approach the task of localisation from an instantaneous observation with no motion or time-coherency cues, and without assuming a good initialisation for pose. Furthermore, unlike the scan-matching methods, we do not require depth information, and in fact we remove the need for any explicit prediction of layout geometry at query time entirely, instead solely leveraging latent representations of layout.

Kim~\etal~\cite{kim2019deep} first proposed the learning of a latent space which captures room layout similarity. Specifically, this is applied to the task of image retrieval, where images are embedded to reflect their underlying room layout and for a given query image the goal is to return other images with similar layouts. Zheng~\etal~\cite{zheng2020structural} later employed a layout embedding space to aid in the prediction of room layouts. However, both of these works focused on room layouts that follow a box approximation, where the layout takes the form of the inside of a convex cuboid. This leads to much reduced variability in the space of possible layouts: there are 11 types of room layout which can be seen in an image, of which~\cite{kim2019deep} only considered 1. We, instead, consider general room layouts, without imposing any assumptions about their structure, leading to more variety in potential layout. However, it is the same concept of a layout latent space on which LaLaLoc's localisation is enabled.

\section{Task and Definitions}

We perform camera pose localisation within a floor plan provided as prior knowledge. Specifically, we localise the 2 DoF camera pose, $P$, to a point in a 2D floor plan. For rendering of the reference layouts, we assume that the camera and ceiling height are given, and produce a 3D floor plan $\mathcal{M}$ of the scene through extrusion of the 2D floor plan. $\mathcal{M}$ consists of only walls, floors and ceilings.

Throughout the paper, we deal with three main data types that may be captured or rendered at a position within the floor plan, $P_p$: $I_p$ is the RGB panorama captured; $L_p$ is a rendered depth image corresponding to $\mathcal{M}$ being projected into the image; $C_p$ is a pointcloud formed by back-projecting $L_p$, in effect a sub-sampling of $\mathcal{M}$.

\section{Latent Layouts}

LaLaLoc consists of a network with two parallel branches to form a quasi-Siamese network. One branch, \imagebranch, computes a feature descriptor for an RGB panorama image, while the other branch, \layoutbranch, computes a descriptor from layout renderings. The embedding for a RGB panorama image should be identical to the embedding computed for a layout rendering at the same camera pose. This means that LaLaLoc is tasked with computing a singular latent representation of the room layout as visible at a particular location within a floor plan, independent of the sensor data used to compute it.

To learn such an embedding space, we take an approach analogous to knowledge distillation~\cite{hinton2015distilling}. Knowledge distillation generally first trains a complex model as the teacher, which is then used to improve the convergence of a less complex model. Instead, we train a model on the easier task: \layoutbranch performing layout-to-layout matching, and use it as a ``teacher" model for a ``student" performing a more complicated task: \imagebranch learning to encapsulate the layout information present in an RGB image. This approach allows us to maximise the richness of our layout latent space by only considering learning the relationships between layouts, before training \imagebranch to also map images to this space. However, this learning formulation in LaLaLoc differs from more general knowledge distillation as we keep \layoutbranch to use for matching with the outputs of \imagebranch. Therefore, the task requires that \imagebranch not only maintains the relative differences between embeddings, and with it their respective layout similarity, but that it is anchored to the corresponding embedding of \layoutbranch.

In the following, we describe the learning of the latent space, as well the mapping of other modalities to it. An overview of this training procedure is shown at the top of Figure~\ref{fig:training} and the architecture used is given in Appendix~\ref{app:arch}. 

\begin{figure}
    \centering
    \includegraphics[width=0.99\linewidth]{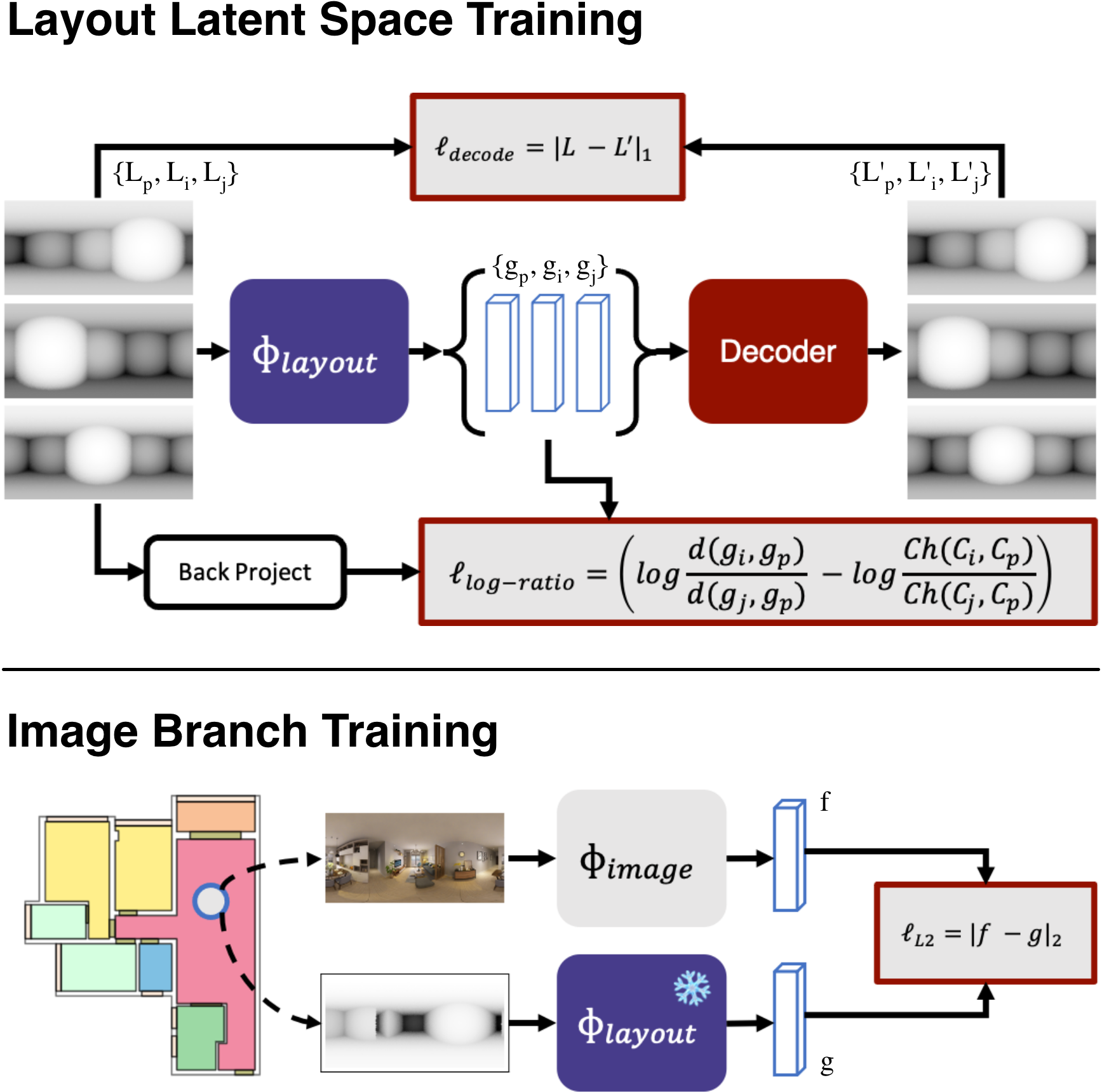}
    \caption{Overview of the LaLaLoc training process. \textit{Top:} The learning of the layout latent space performed solely by \layoutbranch. \textit{Bottom:} The routine used to train \imagebranch to map RGB panoramas to the latent space learned by \layoutbranch.}
    \label{fig:training}
\end{figure}

\subsection{\layoutbranch: Layout-similarity Latent Space}

We first learn a rich latent representation of room layouts by imposing a metric loss for training the layout branch \layoutbranch alone. The embedding space is conditioned so that the distance between the latent representations of layout renderings should reflect the structural difference in their respective layouts. This is achieved through a log-ratio loss formulation, as proposed by Kim~\etal~\cite{kim2019deep}:
\begin{equation}
\label{eq:log_ratio}
    \ell_{log\_ratio}(p, i, j) = \left( log \dfrac{D(g_i, g_p)}{D(g_j, g_p)} - log \dfrac{\text{Ch}(C_i, C_p)}{\text{Ch}(C_j, C_p)}\right)^2,
\end{equation}
\noindent where $(p, i, j)$ represents a triplet of consisting of an anchor p, given by the pose of the RGB panorama, with i and j as two neighbours of p; $g = \Phi_{layout}(L)$ represents the layout embedding computed from layout rendering $L$; $C$ indicates the pointcloud generated from back-projection of the layout depths; $D(\cdot)$ is the Euclidean distance function and $\text{Ch}(\cdot)$ is the Chamfer distance.

In addition to the layout similarity loss, we also leverage a layout decoder for training. The decoder takes the latent vector $g = \Phi_{layout}(L)$ and decodes it to layout depth image, $L'$ thus forming a layout auto-encoder during training. We apply a decoding loss in the form:
\begin{equation}
    \ell_{decode} = |L' - L|_1,
\end{equation}
\noindent with the L1-norm chosen for its favourable performance for depth estimation~\cite{carvalho2018regression}. 

\subsection{\imagebranch: Learning to Match Images to Layouts}

Once the layout latent space has been learned by the teacher branch, we train our student branch, \imagebranch, to embed RGB images to the same layout embedding space. For this training, we freeze \layoutbranch, therefore the RGB branch is tasked only with mapping images to a fixed layout space and can be aided by the response of \layoutbranch. 

We do this simply by applying a loss on the Euclidean distance between the RGB embedding, $f$, and the layout embedding, $g$. This mirrors the localisation strategy where image-layout matching will be predicted through the Euclidean distance between the respective embeddings in the latent space. The loss is given simply by:
\begin{equation}
    \ell_{L2}(p) = |f_p - g_p|_2.
\end{equation}

\section{Localisation}
\label{sec:loc}

While it would be possible to fine-tune the latent representations computed by LaLaLoc when localising in new scenes, we instead propose the approach as a fixed network, which is able to generalise to new scenes without any finetuning or RGB training trajectories. 

LaLaLoc is a 2-stage localisation approach: a coarse, global retrieval stage, followed by cross-modal pose optimisation through differentiable rendering. An overview of the localisation method is depicted in Figure~\ref{fig:teaser}. In the following, we describe each of these stages in more detail.

\subsection{Coarse retrieval}
The coarse retrieval stage operates by sampling candidate poses in a uniform grid across the known floor plan.
At each sampled pose, we render the layout from the known floor plan geometry, which is then used to compute a latent vector. These latent vectors and their associated poses form the reference database for localisation. When performing localisation of a query image, we compute the latent vector of the query image and compute its distance with respect to each entry in the reference database. The coarse localisation estimate is given by the pose belonging to the nearest neighbour latent vector.

\subsection{Pose refinement}
The first, coarse stage of the LaLaLoc localisation method is limited by the density of the sampled poses within the floor plan. We, therefore, include pose refinement to enable more fine estimate of the camera location. We propose two methods for this: a retrieval-based approach which re-samples more densely around the coarse estimate and again localises by retrieval; or direct optimisation of pose by layout similarity in the latent space. 

To more densely explore the region around the nearest neighbour from coarse retrieval, we sample poses in a Vogel Disc~\cite{vogel1979better} centred around the nearest neighbour. This produces an (approximately) even sampling in a circular region around the neighbour. The refined pose is then returned as the most similar from these newly sampled poses. 

We will describe the formulation of the direct optimisation more thoroughly below.

\subsubsection{Latent Optimisation of Pose}
We propose a direct pose optimisation through differentiable rendering. While differentiable rendering-based approaches have been shown to be effective for pose estimation~\cite{palazzi2018end, chen2019learning}, these works rely on homogeneous data to compute losses between the prediction and the target, often employing pixelwise losses based on photo-metric or depth reconstruction errors. However, in our application we again must tackle the challenge of the asymmetry of our query (RGB) and reference (layouts) data types.

Instead, we optimise for pose with latent losses, using distance between embeddings computed by our network to model the matching energy between RGB images and layouts. In doing, so we are able to bridge the gap between data modalities, without the need for explicit prediction. This is achieved by employing a differentiable renderer to compute the layout at a pose estimate, $P_r$, we can ensure that the chain of operations to compute a layout embedding at $P_r$ are differentiable: $g_r = \Phi_{layout}( \Omega(P_r, \mathcal{M})  ),$
where $\Omega(P_r, \mathcal{M})$ is the layout rendering. We therefore can refine the pose using a gradient-based optimisation with objective:
\begin{equation}
    \min_{P_r} D( \Phi_{layout}( \Omega(P_r, \mathcal{M}) ), f_p).
\end{equation}

\section{Dataset}
Training and evaluation is performed on the Structured3D dataset~\cite{zheng2019structured3d}. The dataset consists of 3,500 synthetic indoor scenes. Each scene comprises of multiple rooms, leading to 21,835 rooms in total. Importantly for our task, a 3D floor plan is provided, as well as a photorealistic panorama image rendered for each room in the scene. 

We follow the predefined split of scenes, with 3000 used for training, 250 used for validation and 250 used for testing. We found that some scenes had corrupted data, which were excluded resulting in 2979/246/249 scenes for training/validation/testing. Due to the split being conducted over scenes, it means that all evaluation within this paper is conducted on an unvisited room, where we define unvisited as a scene with no prior image capture \textit{e.g.} a training trajectory captured in the target scene.

For each image, there are three furniture configurations: empty, simple and full. Each of these are rendered in three lighting conditions: warm, cool and raw. Unless otherwise stated, a configuration is selected randomly at each iteration during training and evaluation is performed in the ``full" and ``warm" furniture and lighting settings, respectively -- with full being the most difficult due to the presence of more distractors when inferring layout.

\begin{table*}[]
    \centering
    \begin{tabular}{cc c cc cccc}
    \toprule
        Query & & Layout &  Pose & \multicolumn{5}{c}{Localisation Accuracy}\\
        Type & Model & R@1 & R@1 & Median (cm) & $<$1cm & $<$5cm & $<$10cm & $<$1m \\
        \midrule
         & \textit{Oracle} & \textit{100\%} & \textit{91.1\%} & \textit{20.4} & \textit{0.3\%} & \textit{3.61\%} & \textit{13.5\%} & \textit{92.9\%} \\ 

        \addlinespace[0.5em]
        
        \multirow{2}{*}{Layout} & 2D ICP & - & - & 0.5 & \textbf{87.9\%} & \textbf{99.8\%} & \textbf{99.8\%} & \textbf{99.9\%} \\

        & LaLaLoc & 90.1\% & 87.4\% & \textbf{0.2} & 69.1\% & 82.4\% & 91.4\% & 94.9\% \\ 
        
        \addlinespace[0.5em]
        
        \multirow{3}{*}{RGB/Furnished} & 2D ICP & - & - & 21.8 & \textbf{9.5\%} & 26.4\% & 35.6\% & 68.5\% \\

        & HorizonNet~\cite{sun2019horizonnet} + Loc & 72.4\% & 68.0\% & 9.1 & 3.3\% & 29.3\% & 53.4\% & 77.4\% \\
        
        & LaLaLoc & 72.4\% & 70.6\% & \textbf{8.3} & 3.6\% & \textbf{32.0\%} & \textbf{58.0\%} & \textbf{87.5\%} \\
    \bottomrule
    \end{tabular}
    \caption{Layout localisation accuracy for our trained models and baselines on the Structured3D test set.}
    \label{tab:localisation}
\end{table*}

\begin{figure}[]
    \centering
    \includegraphics[width=0.90\linewidth]{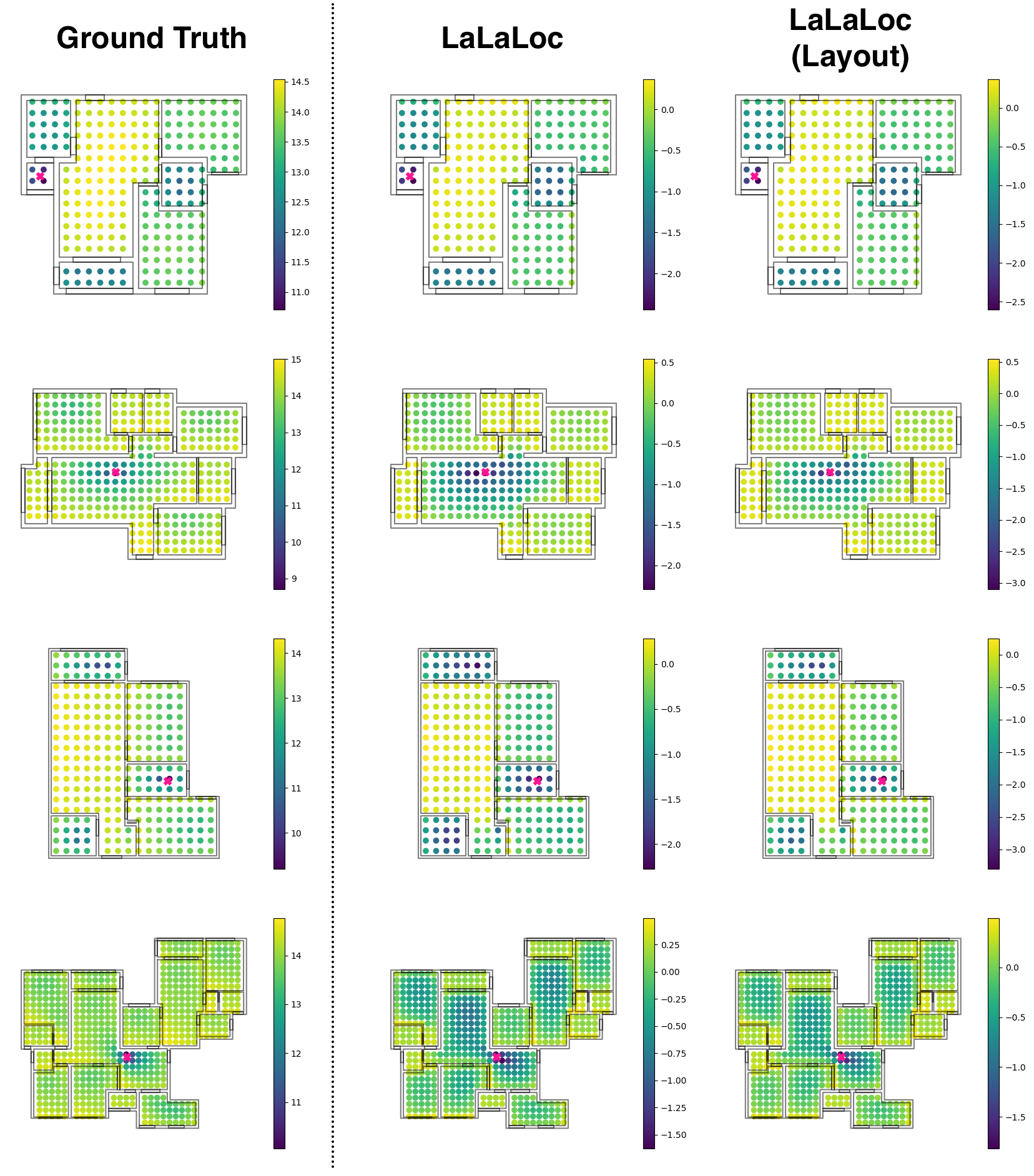}
    \caption{Qualitative depiction of the layout distances between a query pose, marked with a pink ``X'', and the grid of sampled poses across the scene. For each sampled pose, we colour it by the log of its respective distance from the query. \textit{Left:} we plot the ground-truth layout distances. \textit{Middle:} the layout distance is that predicted by LaLaLoc. \textit{Right:} predicted by LaLaLoc in its layout-to-layout configuration.}
    \label{fig:distances}
\end{figure}

\section{Experiments}

We detail the localisation performance of LaLaLoc for our main task of localisation in an unseen environment, as well as analyse LaLaLoc's components and their contribution to the final accuracy. 
Unless otherwise stated, we sample a 2-dimensional grid of locations at a resolution of 0.5m $\times$ 0.5m. We then either take the nearest neighbour pose as our prediction, or use it to initialise the refinement stage. Our Vogel Disc re-sampling is set to a radius of twice the grid resolution. It is worth noting that all evaluations are performed with only the top 1 retrieval, which is the hardest test setting as there is limited scope to recover from a bad retrieval.

We include results for the two main modes of operation for LaLaLoc. The first assumes that the ground truth room layout is known for the query image. In this configuration, the layout branch, \layoutbranch, is used alone. This setting is ideal for determining the expressiveness of the learned layout latent space as there are no discrepancies caused by translation from image to the layout latent space.  The other configuration, however, is the more realistic scenario, where RGB panoramas are used as the query image, with \imagebranch mapping them to LaLaLoc's embedding space.

For a baseline comparison, we include results for an ICP-based method. Similar methods are widely used in fields like robotics for motion estimation~\cite{martinez2006mobile}, scene reconstruction, map building and localisation~\cite{ruiz2017multiversal}. The implementation details of this method are provided in Appendix~\ref{app:icp}. In addition, we compare against a baseline formed from HorizonNet~\cite{sun2019horizonnet}. Given a query image, we use HorizonNet to predict its layout explicitly. We then retrieve the nearest neighbour from the sampled grid via L1 distances between depths. We perform VDR initialised at the nearest neighbour. Finally, we perform a gradient-based optimisation between the predicted layout and layouts rendered from the floor plan. It is worth noting that this is a significant extension to HorizonNet to perform localisation.

Various performance metrics are compiled across the following experiments that pertain to retrieval, and the final estimation of pose. The retrieval metrics provide insight into the performance of the first localisation stage: \textit{Layout Recall @1} measures the proportion of the time the predicted nearest neighbour layout is the most similar layout; \textit{Pose Recall @1} measures the fraction of predicted nearest neighbours that are the neighbour with the nearest pose to the query. Localisation accuracy is given by: \textit{Median Pose Accuracy} as the median distance between query and predicted pose; \textit{Accuracy}$<\tau$ lists the fraction of frames localised to within the threshold, $\tau$.

\subsection{Localisation with a Floor Plan Prior}

Through the full localisation procedure outlined in Section~\ref{sec:loc}, we detail LaLaLoc's performance in Table~\ref{tab:localisation}. Worth considering first is the retrieval accuracy. The layout-to-layout retrieval performance deteriorates only slightly from the layout oracle, suggesting that the latent space is able to capture layout extremely well. 
For qualitative confirmation of this, we plot ground truth and inferred layout distances in Figure~\ref{fig:distances}. There it can be seen that the inferred difference or similarity between layouts is highly representative of the ground truth. Given the increased difficulty of inferring layout from RGB images, the cross modal retrieval performance of LaLaLoc, as seen qualitatively in Figure~\ref{fig:distances}, is very good. This is emphasised by the retrieval accuracy where 87.5\% of retrievals were within 1m of the true pose and thus within the radius of the Vogel Disc refinement.

In the layout-to-layout configuration, LaLaLoc is able to perform competitively against our ICP baseline. This is the ideal case for ICP, since it computes alignment on the pointclouds directly, both of which are known to the exact scale, whereas LaLaLoc introduces a layer of abstraction between the reference and the query. In the panorama-to-layout setting we see that LaLaLoc outperforms the baselines tested, including the method involving explicit layout prediction with HorizonNet~\cite{sun2019horizonnet} and aligning it to the known floorplan geometry. 

\begin{figure}
    \centering
    \includegraphics[width=0.99\linewidth]{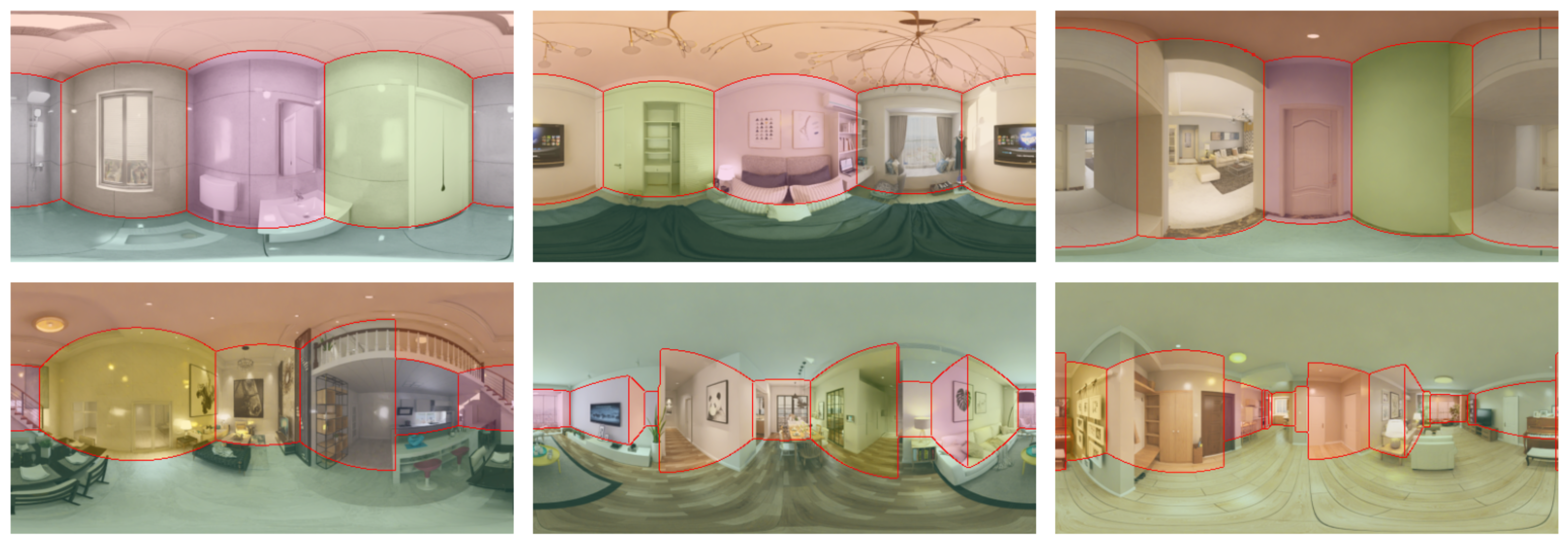}
    \caption{Example failures. \textit{Top:} Cases where the wrong room has been retrieved due to ambiguity of layout, but the alignment is generally coherent with the layout seen in the query image. \textit{Bottom:} The correct room is retrieved, but the alignment is incorrect. In these cases, it is often seen that room corners are incorrectly aligned to the edges of objects such as tables and counter tops.}
    \label{fig:failure_cases}
\end{figure}

We visualise LaLaLoc's two main types of failure in Figure~\ref{fig:failure_cases}. The most common is that an incorrect room with similar geometry is retrieved. In this scenario, the pose is still generally refined to a location that produces a plausible alignment. In the other scenario, the correct room is retrieved but the aligment is incorrect. Often in this scenario we see that floor or wall edges are aligned to objects in the room, suggesting that the refinement may have been caught in a local minimum.

\begin{table}[]
    \centering
    \begin{tabular}{ccccccc}
    \toprule
         & & \multicolumn{4}{c}{Localisation Accuracy}\\
         & Method & Med. & $<$1cm & $<$10cm & $<$1m \\
        \midrule
         & \textit{Retr. Oracle} & \textit{42.7} & \textit{0.0\%} & \textit{2.9\%} & \textit{75.7\%} \\ \addlinespace[0.25em]
        \multirow{2}{*}{\rotatebox[origin=c]{90}{Lay.}} & 2D ICP & 0.7 & \textbf{76.9\%} & \textbf{99.4\%} & \textbf{99.6\%}\\
        & LaLaLoc & \textbf{0.5} & 56.1\% & 70.5\% & 84.3\% \\ \addlinespace[0.25em]
        \multirow{2}{*}{\rotatebox[origin=c]{90}{Pano.}} & 2D ICP  & 21.7 & \textbf{7.3\%} & 34.7\% & 69.4\%\\
        & LaLaLoc & \textbf{11.5} & 2.6\% & \textbf{46.1\%} & \textbf{79.5\%}\\
    \bottomrule
    \end{tabular}
    \caption{Localisation performance when sampling on a lower resolution, 1m $\times$ 1m grid.}
    \label{tab:1m_refinement}
\end{table}

In Table~\ref{tab:1m_refinement}, we provide further results for localisation performance with a lower resolution grid of poses used for sampling. Specifically we sample at a resolution of 1m $\times$ 1m, as opposed to the original 0.5m $\times$ 0.5m.
Despite the retrieval error being more than double that seen with the 0.5m grid, LaLaLoc is still able to recover from this during its refinement stage with final accuracy only seeing a small degradation in accuracy.

\begin{table}[]
    \centering
    \begin{tabular}{ccccc}
    \toprule
        Furniture &  & Layout & Pose & Median\\
        Level & Method & R@1 & R@1 & Accuracy (cm) \\
        \midrule
        \multirow{2}{*}{Empty} & 2D ICP & - & - & \textbf{2.4}\\
        & LaLaLoc & 76.4\% & 74.0\% & 5.7 \\ \addlinespace[0.25em]
        \multirow{2}{*}{Simple} & 2D ICP & - & - & 10.0\\
        & LaLaLoc & 74.8\% & 73.3\% & \textbf{6.7} \\ \addlinespace[0.25em]
        \multirow{2}{*}{Full} & 2D ICP & - & - & 23.0 \\
        & LaLaLoc & 72.4\% & 70.6\% & \textbf{8.3} \\ 
    \bottomrule
    \end{tabular}
    \caption{Impact of furniture level on localisation performance. LaLaLoc is run with a full refinement scheme (VDR + LPO) for computation of localisation accuracy.}
    \label{tab:furniture}
\end{table}

\subsubsection{Robustness to Level of Furniture}
With the introduction of furniture in the second test setting, the effectiveness of LaLaLoc becomes readily apparent. LaLaLoc now significantly outperforms the ICP baseline in nearly all the metrics evaluated. In Table~\ref{tab:furniture}, we further analyse the impact of the furniture configuration on the final localisation accuracy by evaluating performance across those provided in the Structured3D dataset. Here, it becomes apparent that, although the ICP baseline performs very well when operating on the ground-truth layouts, when the query type becomes increasingly furnished its performance drops remarkably. On the other hand, the results show that LaLaLoc sees very little deterioration as the number of distractors in the scene increase from empty to full. This shows that \imagebranch within LaLaLoc is able to accurately infer room layout from the RGB images, even when the room is cluttered. 

\subsubsection{Pose Refinement}
To evaluate the performance of LaLaLoc's pose refinement, here we compare various strategies. Specifically, we investigate the combination of the Vogel Disc re-sampling and latent pose optimisation to improve upon the coarse pose as predicted by retrieval. We also include an explicit prediction and alignment strategy,``VDR + Decode". This is implemented by keeping the decoder from training and using it to predict the query layout explicitly. Pose is then predicted via a gradient-based optimisation on the L1 distance between the predicted layout and rendered layouts from the floor plan. Results are listed in Table~\ref{tab:refinement}. The VDR and LPO refinement stages of LaLaLoc are both shown to be effective, with each able to improve on the retrieval alone. Our latent pose optimisation applied alone outperforms VDR, but the best localisation performance is given by using the re-sampling to initialise the latent pose optimisation.

\label{sec:lpo}
\begin{table}[]
    \centering
    \begin{tabular}{cccccc}
    \toprule
         Refine & \multicolumn{4}{c}{Localisation Accuracy}\\
         Method & Med. & $<$1cm & $<$5cm & $<$10cm \\
        \midrule
        Retrieval  & 22.5 & 0.3\% & 3.19\% & 11.8\% \\
        \addlinespace[0.25em]
        VDR-only & 11.0 & 0.8\% & 15.1\% & 45.3\% \\
        LPO-only & 10.5 & 2.0\% & 27.4\% & 48.4\% \\
        \addlinespace[0.25em]
        VDR + Decode & 12.2 & 0.5\% & 13.2\% & 39.7\% \\
        VDR + LPO & \textbf{8.3} & \textbf{3.6\%} & \textbf{32.0\%} & \textbf{58.0\%} \\
    \bottomrule
    \end{tabular}
    \caption{Comparison of latent pose optimisation against a render-and-compare method, ``Decode", as well as against no optimisation. All refinement methods are initialised at the pose returned from our Vogel disc re-sampling.}
    \label{tab:refinement}
\end{table}

In comparison to the decode baseline, it is clear from the results that the optimisation within the latent space leads to superior results. In fact, the refined poses through decoding performed worse than the retrieval poses used to initialise them. We believe the discrepancy in refinement accuracy can be explained by layouts being easier to capture layout in the latent space, than they are to articulate, \textit{i.e.} to explicitly predict. The simple depth decoder may not be sufficiently intricate to predict layouts suitable for this refinement. After all, prediction of general room layouts is an activate area of research in itself~\cite{howard2018thinking, sun2019horizonnet, stekovic2020general}.

\begin{figure}
    \centering
    \includegraphics[width=0.6\linewidth]{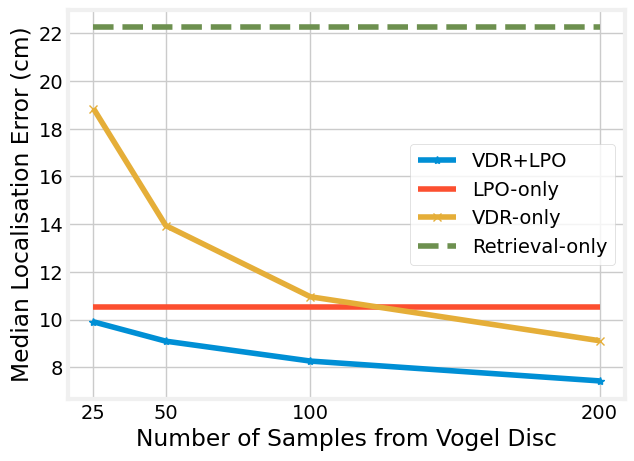}
    \caption{Investigation of the impact of sampling density from the Vogel Disc before performing latent pose optimisation. We include results for performing latent pose optimisation without the Vogel Disc re-sampling beforehand, as well as reference to the initial retrieval accuracy used to initialise these refinement schemes.}
    \label{fig:vdr_sample}
\end{figure}

As a further evaluation of the latent pose optimisation, we perform refinement with a varying number of more densely sampled poses in the naive Vogel Disc refinement strategy. As can be seen in Figure~\ref{fig:vdr_sample}, the benefit from the re-sampling decreases quickly as the number of samples decreases. This experiment also demonstrates the ability of latent pose optimisation to recover from increasingly poor initialisation provided by the Vogel Disc re-sampling. In all tested configurations, the latent pose optimisation was able to improve upon the re-sampled nearest neighbour, and the decrease in accuracy was far less severe.

\subsection{Ablation}

In this section, we validate various components of LaLaLoc's design, from the demonstration that layout localisation is feasible and the metric used for layout similarity, to the training objectives for the network. Unless otherwise stated, the results are all validation accuracy and correspond to retrieval only, with no pose refinement performed. In the experiments we mark our chosen design with an *.

\subsubsection{Layout-similarity Metric}

\begin{table}[]
    \centering
    \begin{tabular}{cccc}
    \toprule
        Similarity & Recall & Pose Error & Correct \\
        Metric & @1 & Median (cm) & Room\\ 
        \midrule
        \textit{Pose} & \textit{100\%} & \textit{19.8} & \textit{100\%} \\ \addlinespace[0.25em]
        Edges & 79.3\% & 20.3 & 83.6\% \\ \addlinespace[0.25em]
        Depth & 80.3\% & 20.1 & 84.0\% \\
        Rel. Depth & 77.1\% & 20.1 & 85.4\% \\ \addlinespace[0.25em]
        Chamfer* & \textbf{90.8\%} & \textbf{20.0} & \textbf{92.1\%} \\
    \bottomrule
    \end{tabular}
    \caption{Evaluation of layout similarity metrics for the localisation task on the Structured3D validation split. \textit{Pose} refers to picking the nearest location in the sampled grid to the query.}
    \label{tab:oracle}
\end{table}

When discussing localisation from a floor plan, it is not unreasonable to think of room shapes identical to one another, thus leading to significant ambiguity and rendering localisation ineffective without other cues. To some degree this ambiguity cannot be completely removed when localising with respect to layout alone. However, we hypothesise that there are significant differences between layout representations and the respective measures of similarity in their susceptibility to this ambiguity. This means that, if the wrong metric is chosen, layout similarity can become a much less expressive signal for localisation. Therefore, we explore which representation of an arbitrary room layout gives the best performance for localisation.

Specifically, we evaluate four representation and layout similarity metric pairs: \textit{Edges}, the 2D Chamfer distance between sets of edge pixels in layouts represented as an edge segmentation, used in~\cite{boniardi2019robot} for floor plan localisation; \textit{Depth}, defined as the L1 distance between layouts rendered as depth images; \textit{Relative Depth}, the L1 distance between relative depth images (max depth value = 1 in all images); and \textit{Chamfer distance}, where the depth images are backprojected and the Chamfer distance is computed between the resulting pointclouds.

We list the results in Table~\ref{tab:oracle}. As can be seen in the table, representing layout as a pointcloud with the Chamfer distance provides the best formulation of those tested, significantly out-performing the image-based similarity metrics tested.

\begin{table}[t]
    \centering
    \begin{tabular}{ccccc}
    \toprule
        Query & Training & Layout & Pose & Median \\
        Type & Routine & R@1 & R@1 & Acc. (cm)\\
        \midrule
        \multirow{2}{*}{\rotatebox[origin=c]{90}{Lay.}} & End-to-End & 81.1\% & 80.4\% & 20.9 \\
        & 2-stage* & \textbf{89.3\%} & \textbf{87.6\%} & \textbf{19.9}\\ \addlinespace[0.25em]
        \multirow{2}{*}{\rotatebox[origin=c]{90}{RGB}} & End-to-End & 24.4\% & 23.7\% & 62.7\\
        & 2-stage* & \textbf{70.7\%} & \textbf{70.5\%} & \textbf{21.7} \\
    \bottomrule
    \end{tabular}
    \caption{Retrieval comparison with an end-to-end training on the Structured3D validation split.}
    \label{tab:e2e}
\end{table}

\subsubsection{End-to-End Training}
We compare our proposed 2-stage training to an end-to-end formulation. We use the same losses as described previously, however we found it best to cut the gradient of $g$ when applied in $\ell_{L2}$. Therefore, \layoutbranch is only optimised to minimise $\ell_{log\_ratio}$ and $\ell_{decode}$, mirroring the 2-stage approach. In Table~\ref{tab:e2e}, we list the results of this training strategy. \layoutbranch is still able to learn a representative latent space, albeit significantly less so than when trained alone. However, \imagebranch performs very poorly when trained in this manner, despite trying a number of different loss scaling factors. We hypothesise that the other losses pollute the gradients to \imagebranch, similar to the introduction of additional losses for training \imagebranch, as explored later. With this, it appears that our 2-stage training is the more effective, or at least significantly more forgiving.

\begin{table}[]
    \centering
    \begin{tabular}{cccc}
    \toprule
        & Layout & Pose & Median \\
        Losses  & R@1 & R@1 & Acc. (cm) \\
        \midrule
        \textit{Oracle} & \textit{100\%} & \textit{90.8\%} & \textit{20.0} \\
        $\ell_{log\_ratio}$ & 78.8\% & 77.3\% & 20.5 \\
        $\ell_{decode}$ & \textbf{90.9\%} & 84.6\% & 20.1 \\
        $\ell_{log\_ratio} + \ell_{decode}$* & 89.3\% & \textbf{87.6\%} & \textbf{19.9} \\
    \bottomrule
    \end{tabular}
    \caption{Layout-to-layout teacher model ablation on the Structured3D validation set. We evaluate the contribution of each of the losses used to train our layout branch.}
    \label{tab:lalalearning}
\end{table}

\subsubsection{Layout Latent Space Learning}

Here we evaluate multiple objective functions to learn LaLaLoc's latent layout embedding space, \textit{i.e.} for training \layoutbranch. Results are listed in Table~\ref{tab:lalalearning}, where we see that the combination of the log-ratio loss and decoder loss provides the most effective latent space for layout-similarity matching and localisation. Notably, it appears that the autoencoder formulation is more representative than training with the relational similarity constraints alone.

\begin{table}[]
    \centering
    \begin{tabular}{ccccc}
    \toprule
         &  & Layout & Pose & Median \\
        Backbone & Predictor & R@1 & R@1 & Acc. (cm)\\
        \midrule
        ResNet18* & FC* & \textbf{89.3\%} & \textbf{87.6\%} & \textbf{19.9}\\
        ResNet50 & FC & 87.5\% & 85.3\% & 20.0\\
        ResNet18 & MLP & 87.2\% & 86.3\% & \textbf{19.9}\\
    \bottomrule
    \end{tabular}
    \caption{Comparison of network architecture for \layoutbranch on the Structured3D validation split.}
    \label{tab:layoutarch}
\end{table}

In Table~\ref{tab:layoutarch} we evaluate some alternative architectures for \layoutbranch. Specifically, we test replacing the ResNet18 backbone with ResNet50, and replacing the single fully connected layer with a 2-layer MLP after pooling. Although all options perform similarly, we find that the simplest architecture produced the best retrieval results.

\begin{table}[]
    \centering
    \begin{tabular}{cccc}
    \toprule
         & Layout & Pose & Median\\
        Losses  & R@1 & R@1 & Acc. (cm)\\
        \midrule
        \textit{Oracle} & \textit{100\%} & \textit{90.8\%} & \textit{20.0} \\
        $\ell_{L2}$* & \textbf{70.7\%} & \textbf{70.5\%} & \textbf{21.7} \\
        $\ell'_{log\_ratio} $ & 59.6\% & 56.9\% & 25.1 \\
        $\ell_{L2} + \ell'_{log\_ratio} $ & 60.7\% & 59.9\% & 24.6 \\
        $\ell_{L2} + \ell_{kd\_lr} $ & 58.9\% & 56.9\% & 25.0 \\
    \bottomrule
    \end{tabular}
    \caption{Image-to-layout model ablation on the Structured3D validation set. We evaluate the contribution of each of the losses used to train \imagebranch from the frozen layout branch.}
    \label{tab:rgblearning}
\end{table}

\subsubsection{Mapping Images to the Latent Space}
\label{sec:img_abl}

When training \imagebranch, there are many possibilities for how to best learn a mapping of RGB images to the existing latent space. In Table~\ref{tab:rgblearning}, we list results of various training objectives considered in the design of LaLaLoc. In this table: $\ell'_{log\_ratio}$ refers to a cross-modal adaptation of the original loss, where $g_p$ is replaced by $f_p$; $\ell_{kd\_lr}$ is a knowledge distillation variant of the log-ratio loss, where the ground-truth layout similarity is replaced by the embedding distances as computed by \layoutbranch. Equations for these losses are provided in Appendix~\ref{app:image_loss}. Interestingly, the most simple objective, reducing the Euclidean distance between the image embedding and its corresponding layout embedding, resulted in the best results.

\section{Conclusion}
In this paper we have presented LaLaLoc, a localisation method which localises RGB queries to a known floor plan by matching in a latent space that encodes layout similarity. We further leverage this expressive latent space for direct pose optimisation through differentiable rendering. We show that LaLaLoc is able to localise with considerable accuracy in unseen or highly dynamic environments.

{\small
\bibliographystyle{ieee_fullname}
\bibliography{egbib}
}
\clearpage
\appendix
\section{Appendix}
\subsection{Implementation Details}

\subsubsection{LaLaLoc Architecture}
\label{app:arch}
We implement \layoutbranch as the convolutional layers from a ResNet18~\cite{he2016deep} network, with the first layer replaced with an equivalent taking a 1-channel input. \imagebranch, on the other hand, is implemented as a ResNet50~\cite{he2016deep} network. In each, we remove the default classifier and replace it with an average pooling layer and a single fully connected layer which takes the embedding dimension to 128d, from 512d or 2048d for ResNet18 and ResNet50, respectively. The embedding vectors are all L2 normalised before any retrieval or comparison task.

\subsubsection{Decoder Architecture}
\label{app:decoder}

The architecture of the decoder that is used to aid training of $\Phi_{layout}$ is defined as follows. First, the latent representation is fed into a fully connected layer which expands it to 2048d. This is reshaped to $(2 \times 4)$ with 256 channels. The up-scaling stage is formed of multiple repeated blocks. Each block comprises of a 2$\times$ bilinear up-sample, a 2D convolution with kernel size 3, a ReLU non-linearity, and a BatchNorm layer. The number of out channels from each block follows the pattern $(128 \xrightarrow{} 64 \xrightarrow{} 32 \xrightarrow{} 32)$.
Finally, depth is predicted with a point-wise convolution which predicts a depth image. With four 2$\times$ up-samples the final resolution of the decoded depth image is $(32 \times 64)$.

\subsubsection{Training Routine}
We train \layoutbranch for a total of 20 epochs, with a batch size of 4. We optimise using SGD, with an initial learning rate of 0.01, and decay by a factor of 0.1 after 10 and 15 epochs. For the log-ratio loss, we follow the positive:negative sampling ratio of 1:20 as used in~\cite{kim2019deep}, therefore a single minibatch contains a total of 84 layouts. We define positives and negatives by their spatial distance from the anchor: positives less than 0.5m from the anchor; negatives greater than 2m from the anchor. 

\imagebranch is trained for a total of 200 epochs, with a batch size of 64. Due to the dearth of RGB images, only a single image-layout pair is sampled per iteration. We again use a SGD optimiser with a learning rate set to 0.1. This is then decayed by a factor of 0.1 after 100 and 150 epochs. \layoutbranch is frozen for the whole training process. Worth noting is that, when \imagebranch is trained with a variant of $\ell_{log\_ratio}$, the training routine follows that of \layoutbranch.

\subsubsection{Image Branch Losses}
\label{app:image_loss}
Here we provide the full equations for the additional image losses evaluated in the ablation in the main paper. First, we restate $\ell_{log\_ratio}$~\cite{kim2019deep}, as used for the layout branch:
\begin{equation}
    \ell_{log\_ratio}(p, i, j) = \left( log \dfrac{D(g_i, g_p)}{D(g_j, g_p)} - log \dfrac{\text{Ch}(C_i, C_p)}{\text{Ch}(C_j, C_p)}\right)^2,
\end{equation}
\noindent where $(p, i, j)$ is a triplet of locations within a floor plan, at which we infer their layouts $(L_p, L_i, L_j)$, $g = \Phi_{layout}(L)$ represents the respective embedding of a layout, $C$ is the back-projection of layout $L$, $D(\cdot)$ is the Euclidean distance and $Ch(\cdot)$ is the Chamfer distance.

For the training of $\Phi_{image}$, we adapt the loss so that the anchor of the triplet now also has an RGB panorama image $(\{I_p, L_p\}, L_i, L_j)$ :
\begin{equation}
    \ell'_{log\_ratio}(p, i, j) = \left( log \dfrac{D(g_i, f_p)}{D(g_j, f_p)} - log \dfrac{\text{Ch}(C_i, C_p)}{\text{Ch}(C_j, C_p)}\right)^2,
\end{equation}
\noindent where the layout embedding $g_p$, has now been replaced with the embedding of the panorama captured in the same location, $f_p = \Phi_{image}(I_p)$. Therefore, this loss aims to ensure $\Phi_{image}$ captures the relative similarities between layouts.

We also include a further adaptation, $\ell_{lr\_kd}$, where we remove ground-truth layout similarities between $(p, i, j)$:
\begin{equation}
    \ell_{lr\_kd}(p, i, j) = \left( log \dfrac{D(g_i, f_p)}{D(g_j, f_p)} - log \dfrac{D(g_i, g_p)}{D(g_j, g_p)}\right)^2,
\end{equation}
such that $\Phi_{image}$ should instead learn to capture the relative similarities between layout embeddings, rather than the layouts themselves. This loss more closely follows typical knowledge discrimination, as the ``student" is trained without ground-truth labels.

\subsection{Vogel Disc Re-sampling}
\begin{figure}
    \centering
    \includegraphics[width=0.5\linewidth]{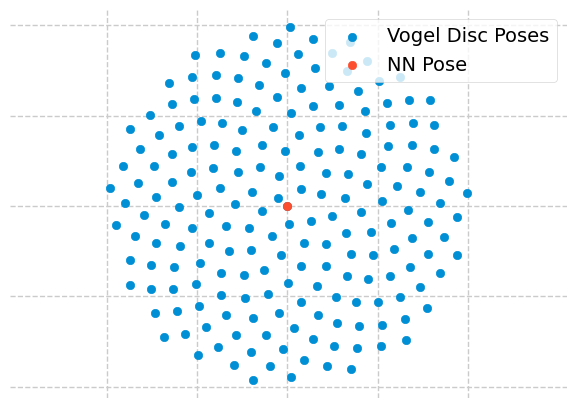}
    \caption{Visualisation of the Vogel Disc sampling pattern with N=200.}
    \label{fig:vogel}
\end{figure}

Our Vogel Disc re-sampling method offers a retrieval-based refinement to the estimated pose from the nearest neighbour in the coarse sampling grid. Specifically, we sample a circular local region centred at the nearest neighbour pose. For $N$ total samples, the $i$th location is given by:
$$
r_i = \sqrt{i / N}\qquad \theta_i = 2\pi (1 - 1 / \phi)
$$
\noindent where $\phi$ is the golden ratio, $r_i$ and $\theta_i$ are the radius and angle in polar coordinates centred at the nearest neighbour pose. The resulting local sampling is visualised in Figure~\ref{fig:vogel}.

\subsubsection{Latent Pose Optimisation}
For our latent pose optimisation, we render layouts using Redner~\cite{li2018differentiable}. We use an Adam optimiser~\cite{kingma2014adam}, with initial learning rate set to 0.01. The learning rate is scaled by a factor of 0.5 as the loss plateaus with a threshold of 0.05 and a patience of 10 iterations. Convergence is considered reached after 20 steps with reduction in the cost less than a threshold of 0.001, or until 150 steps have elapsed.

\subsubsection{Iterative Closest Point Baseline}
\label{app:icp}
In the following, we detail the implementation of our ICP baseline. We emulate a point cloud obtained from a laser scan, $C_m$, through the back-projection of furnished depth images, and perform matching between those and the 2D slices to the point cloud from the known floor plan, $C_f$. At each iteration, the points in $C_m$ are assigned to their nearest counterparts in $C_f$. The rigid transform, $[R,t]$, that aligns $C_m$ and $C_f$ is determined by minimising a point-to-point cost function given this assignment:
\begin{equation}
     E(R,t) = \sum_{i=1}^{|C_f|} \sum_{j=1}^{|C_m|} w_{ij} ||f_i - (Rm_j+t)||
\end{equation}
\noindent where $f_i$ and $m_j$ stand for points in $C_f$ and $C_m$ respectively, being $w_{ij}$ 1 if $f_i$ was matched with $m_j$ and 0 otherwise, and $||\cdot||$ a distance metric (\textit{e.g.} Euclidean distance). The process of assignment and alignment is repeated until a maximum number of 50 iterations is reached, or the update in translation, $t$ is less than a threshold of 0.01mm for 3 successive iterations.

To localise in a room, we initialise the ICP alignment at each of the sampled grid of poses and take the result as the resulting alignment that has the lowest RSME error between point clouds. A typical concern when dealing with this iterative process is its execution time. To handle this, we performed a grid-average sampling method that merges points in the same grid cell, thus preserving the shape of the point clouds but reducing execution time. The grid cell side was heuristically fixed to 5cm.

\subsection{Additional Results}
\subsubsection{Time Complexity}

\begin{table}[t]
    \centering
    \resizebox{\columnwidth}{!}{
    \begin{tabular}{cccccc}
    \toprule
         & \multicolumn{4}{c}{Inference Time (s)}\\
        Method & Retr. & VDR & LPO & Total \\
        \midrule
        2D-ICP & - & - & - & 20.1\\
        LaLaLoc w/o VDR & 0.05 & - & 2.33 (0.85) & 2.38 \\
        LaLaLoc & 0.05 & 2.87 (2.71) & 1.73 (0.63) & 4.65 \\
    \bottomrule
    \end{tabular}
    }
    \caption{Inference time comparison. Times are displayed as: Total time (Render time).}
    \label{tab:time}
\end{table}

In Table~\ref{tab:time} we list the inference times. We omit the rendering time of the sampled grid in the retrieval stage as this time cost is shared between the methods and can be computed offline. In addition, the rendering during VDR could be optimised (currently 2.7s mean). We see that LaLaLoc offers a significant speed advantage over ICP.
Interestingly, VDR appears to reduce the time required for LPO since it provides a better initialisation for LPO, reducing the mean number of iterations to convergence, 41.7 down to 31.4.

\subsubsection{Layout Similarity Low-resolution Sampling}

\begin{table}[]
    \centering
    \begin{tabular}{cccc}
    \toprule
    Similarity & Recall & Pose Accuracy & Correct \\
    Metric & @1 & Median (cm) & Room \\ 
    \midrule
    \textit{Pose} & \textit{100\%} & \textit{40.1} & \textit{99.7\%} \\ \addlinespace[0.25em]
    Edges & 55.6\% & 48.7 & 59.2\% \\ 
    Depth & 55.2\% & 48.3 & 59.5\% \\
    Rel. Depth & 54.1\% & 48.5 & 61.6\% \\ 
    Chamfer & \textbf{71.8\%} & \textbf{42.7} & \textbf{74.4\%} \\
    \bottomrule
    \end{tabular}
    \caption{Evaluation of layout similarity metrics on the Structured3D validation split, now sampled with a lower-resolution 1m $\times$ 1m grid. \textit{Pose} refers to picking the nearest location in the sampled grid to the query.}
    \label{tab:1m_oracle}
\end{table}

Here, we re-perform our evaluation of differing layout representations and similarity metrics for localisation on a lower resolution sampling grid. The results are listed in Table~\ref{tab:1m_oracle}. As can be seen, the lower resolution sampling only serves to exaggerate the performance improvement seen by using the 3D Chamfer distance similarity metric, further emphasising the need for careful selection.

\end{document}